%% file: iclr2021_workshop_after.tex
\documentclass{article} % For LaTeX2e
\usepackage{iclr2021_conference,times}

\input{math_commands.tex}

\usepackage{hyperref}
\usepackage{url}
\usepackage[pdftex]{graphicx}

\usepackage{mathrsfs}
\usepackage{bm}
\usepackage{bbm}
\usepackage{upgreek}
\usepackage{color}
\usepackage{amsthm}
\usepackage{comment}

\newtheorem*{theorem*}{Theorem}
\newtheorem*{corollary*}{Corollary}
\newtheorem*{definition*}{Definition}

\title{Neurons learn slower than they think}

\author{Ilona~Kulikovskikh 
	\\
	Samara University, Russia\\
}

\iclrfinalcopy % Uncomment for camera-ready version, but NOT for submission.
\begin{document}

\maketitle

\begin{abstract}
Recent studies revealed complex convergence dynamics in gradient-based methods, which has been little understood so far. Changing the step size to balance between high convergence rate and small generalization error may not be sufficient: maximizing the test accuracy usually requires a larger learning rate than minimizing the training loss. To explore the dynamic bounds of convergence rate, this study introduces \textit{differential capability} into an optimization process, which measures whether the test accuracy increases as fast as a model approaches the decision boundary in a classification problem. The convergence analysis showed that: 1) a higher convergence rate leads to slower capability growth; 2) a lower convergence rate results in faster capability growth and decay; 3) regulating a convergence rate in either direction reduces differential capability.

\end{abstract}

\section{Introduction}
Training a model is an optimization problem that involves minimizing the model errors on a training dataset. When training a network with gradient-based methods, accelerating convergence to the solution is of a top priority \citep{dieuleveut2017, arora2019}, but not the only performance variable to optimize. Minimizing the difference between the model errors on a training subset and a testing subset, which is called the generalization error, plays a fundamental role \citep{hardt2016, zhan2017, lin2019}.  

While adapting a step size in iterative optimization schemes increases the convergence rate, it does not deliver the best generalization error \citep{luo2019, xie2020, heo2021, zhou2021}. Adaptive optimization methods are often outperformed by non-adaptive stochastic gradient descent (SGD) for overparameterized models, where the number of trainable parameters is much higher than the number of samples they are trained on.
Recent studies revealed that overparametrization itself leads to faster convergence \citep{arora2018, li2018, allen-zhu2019, oymak2019, liu2020, oymak2020, chen2021}. Besides, a step size on testing is usually larger than a step size on training \citep{debortoli2020, li2020, cohen2021}. These findings point to the fact that convergence demonstrates more complex dynamics which has not been well understood so far. 

The study raises a research question on whether we can deepen our understanding by inspecting human and machine reasoning processes in a testing environment. Originating from the item response theory (IRT)~\citep{lord1980, deAyala2009}, \textit{differential capability} shows how fast increases the probability of a correct response to an item of a given difficulty in comparison with a learner's ability.
With a new interpetation in a machine learning context, the differential capability may identify how fast increases the test accuracy compared to a model's ability to reach a decision boundary in a classification problem. It seems to be an appropriate measure to answer our research question. 
%, we can interpret differential capability as a measure which allows analyzing whether the test accuracy increases as fast as the learner reaches a decision boundary in a classification problem.   
%In the item response theory (IRT), \textit{differential capability} shows how fast increases the probability of a correct response to an item of a given difficulty in comparison with a learner's ability. Interpreting this measure in 
%
%
%By introducing \textit{differential capability}, which shows how fast increases the probability of a correct response to an item of a given difficulty in comparison with a learner's ability, into an optimization process, the study explores dynamic bounds of convergence rate. 
%
%These bounds allow balancing between convergence rate and generalization error. 
The work related to this problem is briefly discussed in Appendix \ref{related_work}.
%
%Can we deepen our understanding of convergence dynamics by imitating human behavior in a testing environment? 
%
%Interpreting this in terms of machine behaviour in a testing environment, differential capability allows analyzing whether the test accuracy increases as fast as we approach the decision boundary.
%
%takes into account whether the probability of a correct response to an item of a given difficulty increases as fast as a learner's ability. 
%
%Analyze whether the test accuracy as fast as we 

%% Please note that we have introduced automatic line number generation
%% into the style file for \LaTeXe. This is to help reviewers
%% refer to specific lines of the paper when they make their comments. Please do
%% NOT refer to these line numbers in your paper as they will be removed from the
%% style file for the final version of accepted papers.

\section{Differential capability}

\subsection{Problem setup}
\label{problem}
For a dataset $\{\mathrm{x}_i,y_i\}_{i=1}^m$ with $\mathrm{x}_i\in\mathrm{R}^n$, $y_i\in\{-1,1\}$, let us minimize an empirical loss function with a weight vector $\bm{\uptheta}\in\mathrm{R}^n$:
\begin{equation}
\label{eq::01}
\mathcal{L} (\uptheta) = \sum_{i}\ell (y_i \bm{\uptheta}^\top\mathrm{x}_i),
\end{equation}
where $\ell$ measures the discrepancy between the output $y$ and the model prediction.
The gradient descent (GD) 
%, one of the most dominant first-order optimization algorithms for training neural networks \cite{robbins1951}, 
finds the weight vector with a fixed step size $\eta$:
\begin{equation}
\label{eq::02}
\bm{\uptheta}(t+1) = \bm{\uptheta}(t)-\eta\nabla_{\uptheta}\mathcal{L} (\uptheta). 
\end{equation}

For a large family of monotone losses with polynomial and exponential tails \citep{nacson2019}, the derivative of $\ell(t)$ can be presented as $\ell^{\prime}(t) = -e^{-f(t)}$, where $f(t)$ satisfies $\forall k\in\mathrm{N}$: $\left|\frac{f^{k+1}(t)}{f^{\prime}(t)}\right| = \mathcal{O}(t^{-k})$.
The continuous form of \eqref{eq::02} ($\eta\rightarrow 0$) is equal to $\bm{\uptheta}^\prime (t) = \sum_{i}e^{-f(y_i \mathrm{x}_i^\top\bm{\uptheta}(t))}y_i\mathrm{x}_i$,
%To diminish the influence of step size
%Assume
%$$\bm{\uptheta}' (t) = \sum_{i=1}^m\ell '(\bm{\uptheta}(t)^\mathrm{T}\mathrm{x}_i)\mathrm{x}_i,$$
%where \\
%$\ell '(t) = -a\mathrm{e}^{-f(t)}$, $f(t) = r(t-T)-\left(\frac{E}{r}+\ln_q\left(\frac{N_T}{K}\right)\right)\mathrm{e}^{-rt}+\frac{E}{r}\mathrm{e}^{-rt}$,\\ $a = \varepsilon Kr\left(\ln_q\left(\frac{N_T}{K}\right)+\frac{E}{r}\right)$. 
where the weight vector can be presented asymptotically as $\bm{\uptheta} (t) = g(t)\hat{\bm{\uptheta}}+h(t)$,
$h(t) = o(g(t))$, where $g(t)$ defines a convergence rate, $\hat{\bm{\uptheta}} = \argmin_{\uptheta\in \mathrm{R}^n}\|\bm{\uptheta}\|^2$, so that $y_i\bm{\uptheta}^\top \mathrm{x}_i\geq 1$ \citep{soudry2018, nacson2019} .
Using $\ell^{\prime}(t)$, we can write:
\begin{equation}
\nonumber
g^\prime(t)\hat{\bm{\uptheta}} = \sum_{i}e^{-f(g(t)y_i \mathrm{x}_i^\top\hat{\bm{\uptheta}} + h(t)y_i\mathrm{x}_i^\top} y_i\mathrm{x}_i\\ 
\nonumber \approx
e^{-f(g(t)}\sum_{i}e^{-f^\prime(g(t))h(t)y_i\mathrm{x}_i^\top}y_i\mathrm{x}_i.
\end{equation}
For the last equation, we can require $g^\prime(t) = e^{-f(g(t)}$. Approximating it 
with $g^\prime(t)\approx e^{-f(g(t)) - \ln f^\prime(g(t))}$ gives us a closed from solution $g(t) = f^{-1}(\ln t+C)$.

The present study enriches the provided reasoning with differential capability, which measures whether the test accuracy increases as fast as a model's ability to reach a decision boundary.
Introducing differential capability into an optimization process, this research explores the dynamic bounds of a convergence rate and reveals how an increase/decrease in a convergence rate affects the proposed measure. 
%The smaller differential capability a learner has, the lower ability he/she demonstrates.  

%Introducing the measure into an optimization process, it explores dynamic bounds of a convergence rate and reveals how an increase/decrease in a convergence rate affects the proposed measure.      

Related work often attributed the success in balancing convergence and generalization to the complexity and capacity of neural networks. To observe the impact of differential capability on convergence dynamics, which is not affected by network architecture, the present study focuses on the simplest learner model - a single neuron, the capacity of which stimulates the renewed interest \citep{frei2020, gidon2020, jones2020, yehudai2020}. 

%Simplifying the provided reasoning, we can build a straightforward computational chain ...

\subsection{Loss function with differential capability}
Let us build a loss function on the well-studied two-parameter logistic item response theory (2PL IRT) model ~\cite{lord1980, deAyala2009}: $P(y_{ij} = 1|\omega_j,r_i,d_i) = \frac{1}{1+\exp(-r_i(\omega_j-d_i))}$, 
where $P(y_{ij} = 1|\omega_j,r_i,d_i)$ is a probability of correctly responding $y_{ij} = 1$ to an item $i$ with a difficulty $d_i$ by a learner $j$ with the ability $\omega_j$; $r_i$ is a discrimination parameter that measures the differential capability of an item $i$. A high value of $r_i$ means that the probability of a correct response to an item with a given difficulty increases as quickly as a learner's ability. When $r_i = 1$ and $d_i = 0$, the 2PL IRT model reduces to the sigmoid function. 

%. This means that items located toward the right side have difficulty $\beta$. They require a learner to have greater proficiency $\theta$ to correctly answer items located on the right side than items located on the left side. In general, items located below 0 are ``easy'' while items above 0 are ``hard''.

\begin{figure}[!h]
	\begin{center}
		\includegraphics[width=0.9\textwidth]{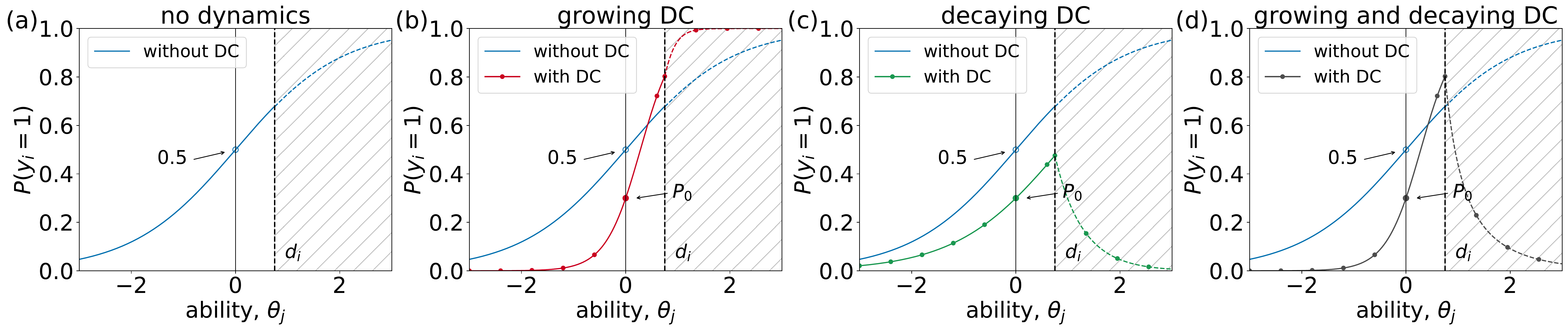}
	\end{center}
	\vspace{-5mm}
	\caption{Different configurations of the DC loss function: (a) no DC ($r_i = 1, c_i = 0$), (b) growing DC ($r_i >0, c_i = 0$), (c) decaying DC ($r_i = 1, c_i >0$), and (d) both growing and decaying DC ($r_i > 0, c_i >0$).}
	\label{dc:func}
	\vspace{-3mm}
\end{figure}

In comparison with the 2PL IRT model, the new loss function, equipped with differential capability, changes the shape of the sigmoid so that it becomes non-monotonic and exhibits more complex behavior. First, differential capability grows with a rate $r_i$. When a learner acquires the ability $\omega_j$ to respond correctly to an item with difficulty $d_i$, the differential capability decays with a rate $c_i$. The probability of answering correctly to an item is equal to $P_{d_i}$. Using the Gompertz logistic law of population dynamics \citep{gray2017}, a learner's response to an item $i$ can be defined as:
\begin{equation}
\label{new_ell}
P(y_{ij} = 1|\omega_j,r_i, a_i, b_i, d_i) = a_ie^{b_ie^{-r_i(\omega_j-d_i)}},
\end{equation}
where $a _i= e^{\varepsilon_i}$, $\varepsilon_i = \frac{c_i}{r_i}$, $b_i = \ln P_{d_i}-\varepsilon_i$. The parameter $\varepsilon_i$ reflects the balance between a growing rate $r_i$ and a decaying rate $c_i$ for an item $i$.
Fig.~\ref{dc:func} depicts different configurations of the DC loss function, where DC stands for differential capability.

\subsection{Convergence analysis}
\label{analysis}
Interpreting differential capability in a machine learning context, where a high value of this measure points out that the test accuracy increases as fast as a model ability to reach a decision boundary, the \eqref{new_ell} can be rewritten as $\ell^{DC}(t) = ae^{be^{-r(t-d)}}$, for which 
$\ell^{\prime DC}(t) = -abre^{-f(t)}$, where $f(t) = r(t-d) - be^{-r(t-d)}$. 
According to the reasonong presented in Section~\ref{problem}, estimating the inverse function of $f(t)$ gives the convergence rate:
\begin{equation}
g^{DC}(z) = d+\left(\mathrm{W_0}(be^{- z})+ z\right)/r,\hspace{2mm}z>0, 
\label{eq:g_f}
\end{equation}
where $\mathrm{W_0}(z)$ is the principal branch of the Lambert function.
\citet{nacson2019, soudry2018} showed that for any strict monotone loss $\ell(t)$, given in Section~\ref{problem}, under certain conditions, $g(t) = \ln t$. With a variable substitute $z = \ln t$, the convergence rate $g(z) = z$ is further refered to as the default rate.

Let us first analyze the parameters $b$, $d$, and $r$, which affect the convergence rate \eqref{eq:g_f}. We can see that the difficulty denoted by $d>0$ increases the absolute value of $g^{DC}(z)$. The parameter $b$ depends on $P_d$ and the ratio $c/r$: $b = \ln P_d-c/r$ (see \eqref{new_ell}). As $0<P_d<1$,  $\ln(P_d)<0$. The smaller $P_d$ is, the faster $|\ln P_d|$ increases. The ratio $c/r>0$ grows up if $r\rightarrow 0$ (an infinitesimal value) or/and $c>r$. The parameter $b<0$, but smaller $P_d$, $r$ and larger $c$ increase its absolute value.

Let us now show that differential capability dynamically changes the bounds of convergence rate.
\begin{theorem*}
	For any $z>0$, $b<0$, moderate $r>0$, and $d = 0$, the bounds of the convergence rate $g^{DC}(z)$ given by \eqref{eq:g_f} are below and above the default convergence rate $g(z) = z$.
\end{theorem*}
The proof of the theorem is  deferred to Appendix \ref{proof}.
\begin{corollary*}
	The bounds of $g^{DC}(z)$ move to the left when $r$ is larger and to the right when $d$ is larger and $r$ is smaller.
\end{corollary*}
\vspace{-4mm}
\begin{proof}
	The validity of the corollary follows from \eqref{eq:g_f}.
\end{proof}

From the convergence analysis, we can conclude that: 1) a higher convergence rate leads to a lower growth rate $r$, which results in smaller differential capability; 2) a lower convergence rate leads to a higher growth rate $r$, which is compensated by a higher decay rate $c$, and, thus, results in smaller differential capability again. This means that regulating a convergence rate in either direction does not increase differential capability.

\section{Experimental results}
\label{others}
\begin{figure}[h]
	\begin{center}
		\includegraphics[width=0.85\textwidth]{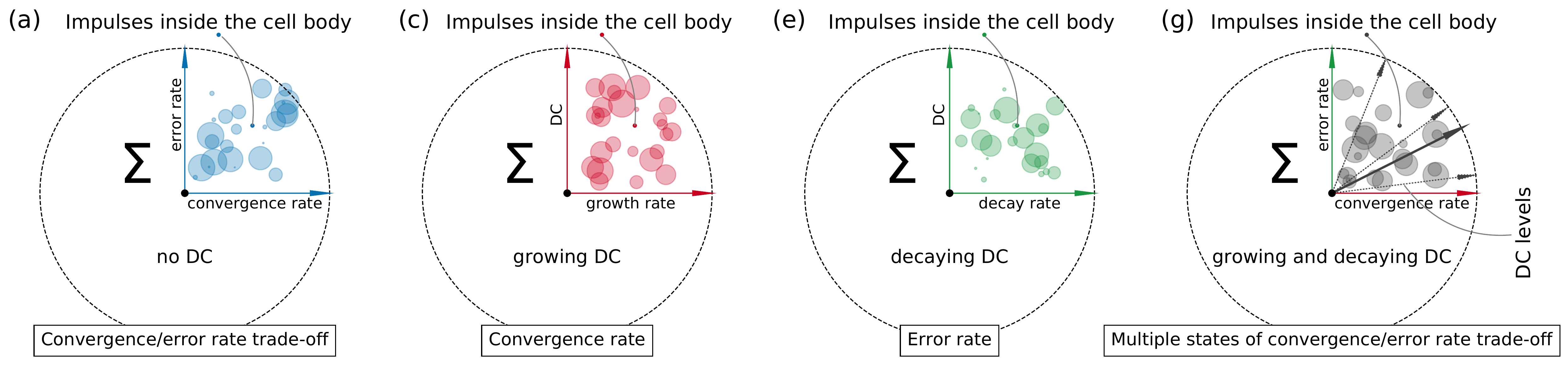}\\
		\includegraphics[width=0.85\textwidth]{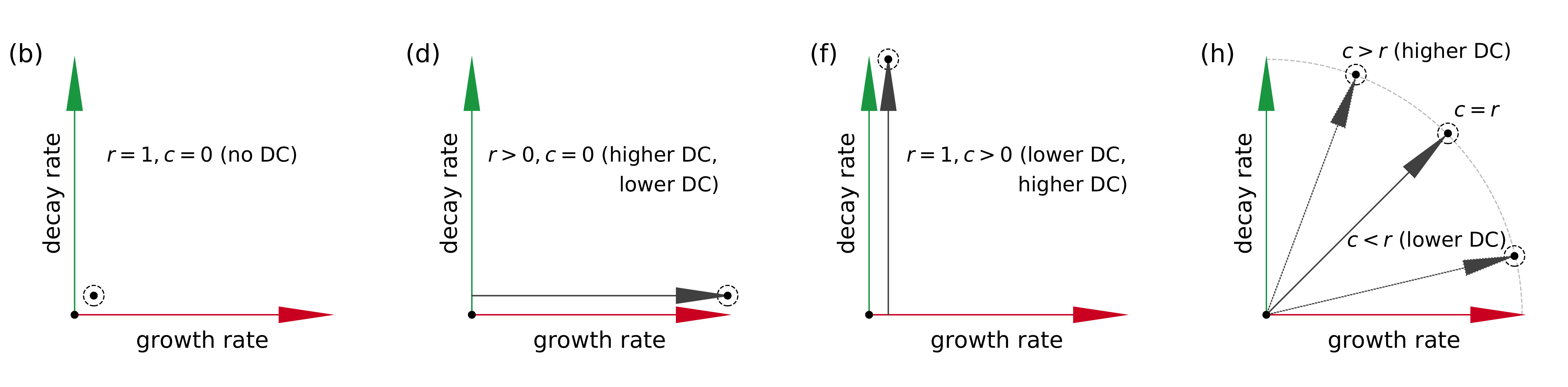}
	\end{center}
	\vspace{-6mm}
	\caption{The impact of differential capability (DC) on inner processes inside a neuron}
	\label{fig:DC_dynamics}
	\vspace{-5mm}
\end{figure}

As the proposed measure dynamically changes the bounds of the convergence rate $g(t)$, it also brings more flexibility in regulating the trade-off between a convergence rate and an error rate.   
Fig.~\ref{fig:DC_dynamics} illustrates how differential capability affects the inner processes inside a neuron for the loss configurations given in Fig.~\ref{dc:func}. It replaces the superposition in \eqref{eq::01} with more complex dynamics (see Fig.~\ref{fig:DC_dynamics} (a), (b) in comparison with Fig. ~\ref{fig:DC_dynamics} (c)-(h)), where the balance between a growth rate $r$ and a decay rate $c$ regulates the convergence/error rate trade-off.

\begin{figure}[h]
	\vspace{-2mm}
	\begin{center}
		\includegraphics[width=0.85\textwidth]{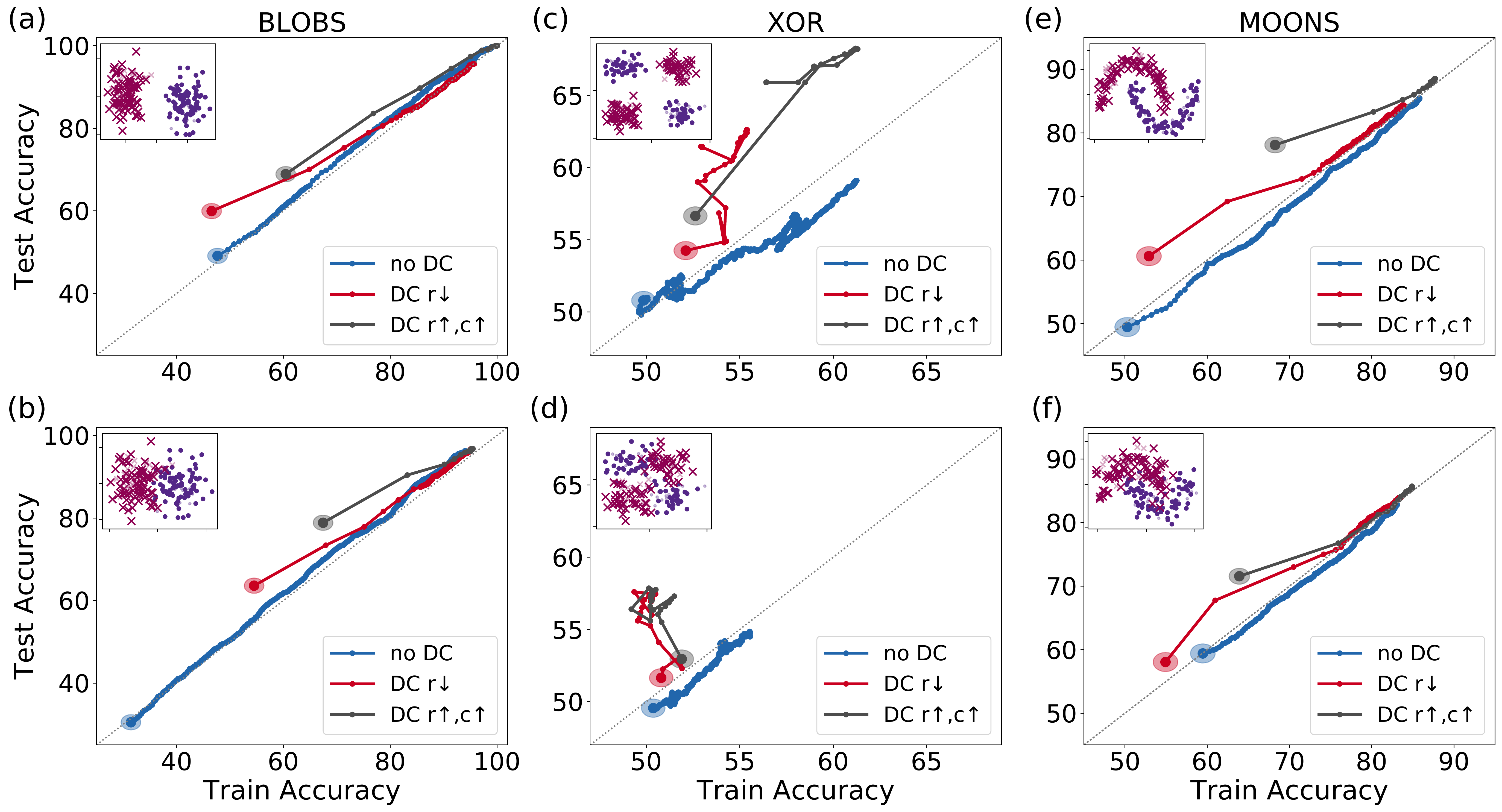}
	\end{center}
	\vspace{-5mm}
	\caption{The impact of differential capability (DC) on convergence/error rate trade-off}
	\label{fig:DC_exps}
	\vspace{-3mm}
\end{figure}

Let us adopt this illustration to design the experiments on a set of synthetic datasets ($m = 1000$, $n = 2$), which were randomly split into training (80\%) and testing (20\%) subsets (see Fig. \ref{fig:DC_exps}, in the upper left corner).
A neuron adjusted its weights with gradient descent in a stochastic setting (SGD) with the default parameters, batch size $|B(t)| = 75$, and $n_\mathrm{epoch} = 1500$. 
The number of runs was equal to 10. The hyperparameters were chosen within the following regions: $d\in[0,5]$, $P_d\in [0.1, 0.9]$, $r\in[0.1,12]$, and $c\in[0,12]$ with a 2.5\% random pick from the full grid space. In Fig. \ref{fig:DC_exps}, the label ``no DC'' reflects the default configuration with the sigmoidal loss function (see \ref{fig:DC_dynamics} (a), (b)), ``DC $r\downarrow$'' denotes the configuration with growing DC \ref{fig:DC_dynamics} (c), (d)), and ``DC $r\uparrow$, $c\uparrow$'' stands for the configuration with growing and decaying DC \ref{fig:DC_dynamics} (g), (h)). Here, the configuration with only decaying DC was left with little  attention. According to the provided convergence analysis, increasing $c$ affects the dynamics bounds of convergence rate to a smaller extent (see Section \ref{analysis}).

The theoretical analysis revealed that a slower growth rate $r\downarrow$ reduces differential capability while increasing the convergence rate. Fig. \ref{fig:DC_exps} illustrates this result. The growth rate $r$ increases slower than it needs to reach the highest test accuracy. But, its value $r>1$, which means it naturally enlarges the step size of the optimizer (see Section \ref{analysis}, $\ell^{\prime DC}(t)$). As a consequence, we can observe the higher test accuracy (see the red curves in contrast to the blue ones on the plots). This is exactly the phenomenon this study is intended to demystify.

A faster growth rate  $r\uparrow$ and decay rate $c\uparrow$ reduce differential capability as well while decreasing the convergence rate. A non-zero value of $c$ balances against even higher value of $r$, which, on the one hand, slows down the convergence, on the other hand, increases the test accuracy to a greater extent (see the black curves in contrast to the red and blue ones on the plots). From the above empirical analysis, we can conclude that neither increase nor decrease in a convergence rate improves differential capability as the model does not achieve the highest test accuracy. 

 %mean +/- std on accuracy (train-test)

%We report average accuracy over 1000 randomly sampled episodes

%We find that our simple ... method outperforms complex ... models.

%\begin{table}[h]
%\caption{Sample table title}
%\label{sample-table}
%\begin{center}
%\begin{tabular}{ll}
%\multicolumn{1}{c}{\bf PART}  &\multicolumn{1}{c}{\bf DESCRIPTION}
%\\ \hline \\
%Dendrite         &Input terminal \\
%Axon             &Output terminal \\
%Soma             &Cell body (contains cell nucleus) \\
%\end{tabular}
%\end{center}
%\end{table}

\section{Conclusion}
This study explored the dynamic bounds of a convergence rate with differential capability, 
which measures how fast increases the test accuracy compared to a model's ability to reach a decision boundary in a classification problem. 
The provided analysis enriched the understanding of convergence dynamics and revealed that both increase and decrease in a convergence rate reduce differential capability.

%\subsubsection*{Author Contributions}
%If you'd like to, you may include  a section for author contributions as is done
%in many journals. This is optional and at the discretion of the authors.

%\subsubsection*{Acknowledgments}
%Use unnumbered third level headings for the acknowledgments. All
%acknowledgments, including those to funding agencies, go at the end of the paper.

\bibliography{iclr2021_workshop}
\bibliographystyle{iclr2021_conference}

\appendix
\section{Related work}
\label{related_work}
The analysis of SGD based optimization for overparameterized models has recently become an active area of research interest \citep{arora2018, li2020, arora2019, allen-zhu2019, zhou2021, wu2021}. Recent studies indicated that large learning rates can preserve good generalization and accelerate SGD convergence with no additional gradient scaling. 
While analyzing the effect of overparametrization, \citet{wu2021} pointed to the difference in directional biases for SGD and GD with a moderate and annealing rates.
\citet{vaswani2019} explored line-search techniques and provided heuristics to automatically set larger learning rates.
\citet{li2020} analyzed an exponential rate schedule. They showed that using SGD with momentum \citep{liu2020} and an exponentially increasing rate, coupled with batch normalization, maintains a good balance between convergence and generalization across all standard architectures. 

In line with more successful SGD adoption for overparameterized models, remarkable progress has been achieved in optimization methods with adaptive learning rates. 
SGDP and AdamP use effective rates without changing the update directions \citep{heo2021}, which allows preserving the original convergence properties of GD optimizers. RAdam adopts the learning rate warm-up heuristic to rectify the variance of adaptive rates \citep{liu2020} and, by that, stabilize training, accelerate convergence, and improve generalization. 
To balance generalization and convergence on unstable and extreme learning rates, \citet{luo2019} put forward AdaBound and AMSBound which adopt dynamic bounds on rates to eliminate the generalization gap between adaptive methods and SGD and maintain a higher learning rate early in the training. These methods were further developed with regard to a dynamic decay rate in \citep{liang2020}.

Adopting the item response theory to address interpretability and explainability issues in deep learning has been reported to be successful. \citet{kulikovskikh2017} extended the model of logistic regression with 4PL IRT model to reduce the disruptive influence of floor and ceiling effects on the convergence of log-likelihood. \citet{lalor2018} inverstigated the relationship between items difficulty and model performance in deep networks. \citet{martinez2019} interpretated the IRT model parameters in terms of a classification problem. \citet{chen2019} proposed a new IRT model,  which allows simulating continuous responses and enriches the family of Item Characteristic Curves. The authors applied the model to evaluating the quality of different machine learning classifiers with items difficulty and discrimination. \citet{kulikovskikh2020} suggested a new query strategy for an active learning environment to increase the transparency of deep network architectures.

\section{Proof of the theorem}
\label{proof}
\begin{proof}
	%Let us now analyze each expression separately.
	For $z>0$, the equation $we^w = z$ 
	has one positive solution $w=\mathrm{W_0}(z)$, which increases with $z$. If $z=e$, then $w = 1$. Thus, $w>1$ if $z>e$. By taking logarithms of both sides, we get:  
	\begin{align}
	\nonumber
	\ln w + w & = \ln z;\\
	\label{link:1}
	w & = \ln z - \ln w < \ln z.
	\end{align}
	When $z>e$, 
	\begin{align}
	\nonumber
	1 & < w < \ln x\\
	\label{link:2}
	0  & < \ln w < \ln\ln z. 
	\end{align}
	
	Substituting \eqref{link:2} into \eqref{link:1} yields:
	\begin{equation}
	\ln z - \ln\ln z  < w < \ln z,
	\end{equation}
	where the left side is positive for $z > 1$.
	Since $w = \mathrm{W_0}(z)$, we can write:
	\begin{equation}
	\ln z - \ln\ln z  < \mathrm{W_0}(z) < \ln z,
	\end{equation}
	
	Let us now modify the argument of $\mathrm{W_0}(z)$ with regard to $g^{DC}(z)$:
	\begin{align}
	\nonumber
	\frac{b}{z} + z < \mathrm{W_0}(be^{-z}) + z &< \frac{b}{z} - \frac{b}{\ln z} + z;\\
	\nonumber
	\frac{b}{z} + z < \mathrm{W_0}(be^{-z}) + z &< b\frac{\ln z-z}{z\ln z} + z,
	\end{align}
	where $\ln z - z < 0$ as $\ln z < z$ for all $z > 0$.
	
	By definition, $b < 0$. Thus, for $z > e$: 
	\begin{align}
	\nonumber
	b\frac{\ln z - z}{z\ln z} + z &> z;\\
	\nonumber
	\frac{b}{z} + z &< z
	\end{align}
	As we see, the boundaries of $\mathrm{W_0}(be^{-z}) + z$ are below and above the default convergence rate $z$. 
\end{proof}

%\begin{proof}
%	%Let us now analyze each expression separately.
%	For $z>0$, the equation $we^w = z$
%	has one positive solution $w=\mathrm{W_0}(z)$, which increases with $z$. If $z=e$, then $w = 1$. Thus, $w>1$ if $z>e$. By taking logarithms of both sides, we get:  
%	\begin{equation}
%	\ln w + w  = \ln z; \hspace{2mm} w  = \ln z - \ln w < \ln z.
%	\label{link:1}
%	\end{equation}
%	When $z>e$, $1 < w < \ln x$; $0  < \ln w < \ln\ln z$
%	%	\begin{equation}
%	%	1 < w < \ln x; 0  \hspace{2mm}  < \ln w < \ln\ln z. 
%	%	\label{link:2}
%	%	\end{equation}
%	Substituting this into \eqref{link:1} yields $\ln z - \ln\ln z  < w < \ln z$, 
%	where the left side is positive for $z > 1$.
%	Since $w = \mathrm{W_0}(z)$, we can write: $\ln z - \ln\ln z  < \mathrm{W_0}(z) < \ln z$.
%	Let us now modify the argument of $\mathrm{W_0}(z)$ with regard to $g(z)$:
%	\begin{equation}
%	\frac{b}{z} + z < \mathrm{W_0}(be^{-z}) + z < \frac{b}{z} - \frac{b}{\ln z} + z;
%	\hspace{2mm}
%	\frac{b}{z} + z < \mathrm{W_0}(be^{-z}) + z < b\frac{\ln z-z}{z\ln z} + z,
%	\nonumber
%	\end{equation}
%	where $\ln z - z < 0$ as $\ln z < z$ for all $z > 0$.
%	By definition, $b < 0$. Thus, for $z > e$: $b\frac{\ln z - z}{z\ln z} + z > z$ and $\frac{b}{z} + z < z$.
%	
%	As we see, the boundaries of $\mathrm{W_0}(be^{-z}) + z$ are both below and above the default convergence rate $z$. 
%\end{proof}

\end{document}

%% file: math_commands.tex
\usepackage{amsmath,amsfonts,bm}

\def\eqref#1{equation~\ref{#1}}
\def\1{\bm{1}}

\DeclareMathAlphabet{\mathsfit}{\encodingdefault}{\sfdefault}{m}{sl}
\SetMathAlphabet{\mathsfit}{bold}{\encodingdefault}{\sfdefault}{bx}{n}

\DeclareMathOperator*{\argmin}{arg\,min}